# Thermal Infrared Colorization via Conditional Generative Adversarial Network

Xiaodong Kuang   Xiubao Sui   Chengwei Liu   Yuan Liu   Qian Chen   Guohua Gu
Nanjing University of Science and Technology, Nanjing 210094, China

## Abstract

*Transforming a thermal infrared image into a realistic RGB image is a challenging task. In this paper we propose a deep learning method to bridge this gap. We propose learning the transformation mapping using a coarse-to-fine generator that preserves the details. Since the standard mean squared loss cannot penalize the distance between colorized and ground truth images well, we propose a composite loss function that combines content, adversarial, perceptual and total variation losses. The content loss is used to recover global image information while the latter three losses are used to synthesize local realistic textures. Quantitative and qualitative experiments demonstrate that our approach significantly outperforms existing approaches.*

## 1. Introduction

Due to the lowering of prices and the increasing of resolution, thermal infrared cameras have gradually migrated from the original military use to many fields [1], such as medicine, security and archeology. People can use thermal infrared cameras to see invisible heat radiation emitted or reflected by all objects regardless of lighting conditions. Since people gradually increase their focus on the application of thermal cameras, thermal infrared colorization, which can improve viewing performance and reduce reaction time, is becoming more and more important.

Colorization of grayscale visible images is an ambiguous problem: is an apple red or green? This has no unique solution because there may be multiple color values for one grayscale value. Despite this ambiguity, current approaches [2-10] have achieved impressive results.

Unlike colorization of grayscale visible images that only estimates the chrominance, we need to simultaneously estimate luminance and chrominance for thermal infrared images. In addition, objective appearance (thermal signature) in thermal infrared images has no necessary relation with its visible appearance (perceived color), making it more difficult for generating realistic colors. So colorizing thermal infrared images can be considered as synthesizing RGB images from semantic maps of thermal infrared images.

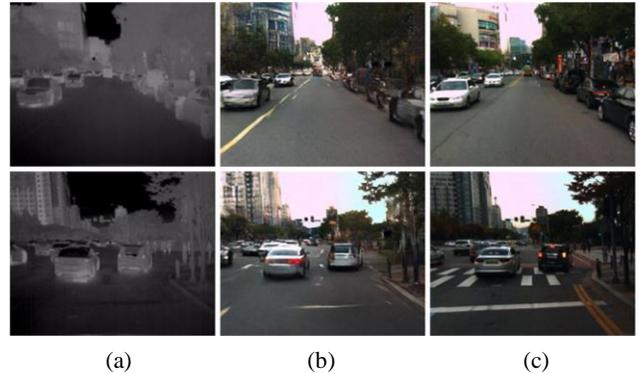

Figure 1: (a) Thermal infrared traffic images. (b) Colorized results by our method. (c) True RGB images.

In this paper, we address thermal infrared colorization problem using a conditional generative adversarial network (TIC-CGAN). We train our model on a dataset containing a large number of thermal infrared and color image pairs. We build on the success of deep learning and Generative Adversarial Networks (GANs) [11]. While Berg *et al.* [12] have already proposed an autoencoder architecture based on convolutional neural network (CNN), it is still very crucial to perform a refined training procedure. RGB images generated using their model have blurry details and lack textures. To address this issue, we use a conditional GAN-based model and define a new objective function. Our generator is a coarse-to-fine generator designed for high-resolution image synthesis [13], while our discriminator is a $70 \times 70$ PatchGAN derived from image-to-image translation [14]. Our objective function combines content, adversarial, perceptual and total variation (TV) losses to achieve realistic image quality. Content loss aims to generate global colors and adversarial loss helps generate local details. Both perceptual and TV losses make the local details more fine. As shown in Fig. 1, using both the conditional GAN-based architecture and multi-term loss function is critical for obtaining realistic colorized results.

Overall, our main contributions include:
- a novel method for thermal infrared image colorization task based on learning a mapping function between thermal infrared and RGB images.
- a multi-term objection function that combines content, adversarial, perceptual and total variation losses, encouraging an appealing image quality generation.
- extensive experiments on the publicly available dataset using TIC-CGAN, demonstrating its superiority over existing methods.

## 2. Related work

### 2.1. Existing colorized methods

Previous visible grayscale colorization works utilize user input and guidance, including scribble [8], label [15], feature extraction [16], and similar images [5, 17-18]. Now the successful applications of CNNs have encouraged people to use deep learning-based approaches for visible grayscale [2-4, 6, 9, 10, 19], near infrared [20-22], and thermal infrared [12] colorization tasks.

Zhang *et al.* [10] propose a CNN-based automatic method to produce realistic colors for visible grayscale colorization. To address the multimodal nature of colors they introduce a classifier and adopt class-rebalancing during training phase. Larsson *et al.* [7] predict per-pixel histograms to address the issue of diverse colorization. Further, Cao *et al.* [2] employ a conditional GAN model and Deshpande *et al.* [3] employ a variational autoencoder.

Near infrared energy is the portion of the electromagnetic spectrum just past the red segment of visible light. Therefore, near infrared and visible light are very close, especially for a red channel of the RGB image. Limmer *et al.* [20] present a deep multi-scale CNN for near infrared colorization. To preserve details, the high frequency features are transferred to the predicted RGB images. Suarez *et al.* [21] also propose a deep conditional GAN model for colorizing near infrared images. Compared to [22], they use a triplet network to estimate each channel separately.

Colorization approaches mentioned above predict the chrominance from the luminance, and therefore cannot be directly applied to thermal infrared images, where the luminance also needs to be predicted. Unlike near infrared, thermal infrared is emitted energy that is sensed digitally, whereas the near infrared (also called the photographic infrared) is reflected energy. So the thermal infrared images reflect the temperature information of an object rather than the color information, which meaning that the color of the object can only be predicted by some higher level semantic information. Recently Berg *et al.* [12] present two fully automatic thermal infrared to RGB image colorization approaches. Their method is robust to image pair misalignments. However, the details in their colorized results are blurred and the sky has artifacts.

### 2.2. Generative adversarial networks

Generative adversarial networks (GANs) have shown excellent performance in a wide range of computer vision tasks such as image synthesis [13], image super-resolution [23], image-to-image translation [14], photo enhancement [24, 25], and image style transfer [26, 27, 28]. A regular GAN architecture is composed of two neural networks: a generative network and a discriminative network. The generative network is used to generate real-looking images while the discriminative network tries to identify which one is a fake. Our method also utilizes this GAN-based architecture to generate realistic RGB images with more fine details.

### 2.3. Image-to-image translation

Recent works have utilized adversarial training for image-to-image translation [14]. Compared to $L_1$ loss, which often produces blurry details [14, 29], the adversarial loss has dominated various image-to-image tasks [23, 30-34]. It can automatically adopt to the difference between the generated and real images. For instance, pix2pix [14] uses conditional GANs for many applications, like photo generation and semantic segmentation. Various unpaired image-to-image translation frameworks have also been proposed in the absence of paired data [35-37].

Recently, Wang *et al.* [13] propose a coarse-to-fine generator to synthesize high-resolution images from semantic label maps using conditional GANs. Their model can output images with realistic textures and fine details. The proposed approach is motivated by their success. We leverage their coarse-to-fine generator architecture and propose a new objective function. Results show that our method yields visually more realistic RGB images than Berg's [12] approach. Side-by-side comparisons clearly confirm our advantage (see Fig. 3).

## 3. Proposed method

Our work aims to learn a mapping from thermal infrared image $x$ to color RGB image $y$. As illustrated in Fig. 2, our network consists of a generator $G$ and a discriminator $D$. $G$ is trained to transform a thermal infrared image into a color RGB image, while $D$ assists $G$ to generate more realistic results. To penalize the distance between colorized and ground-truth RGB images, a content loss based on an L1 term is defined. An adversarial discriminator, a perceptual loss and TV complete our loss definition. The discriminator aims to synthesize fine details while the perceptual and TV losses are combined to improve the quality of details. Therefore, our objective comprises: i) content loss, ii) adversarial loss, iii) perceptual loss, and TV loss. We now detail each loss term.

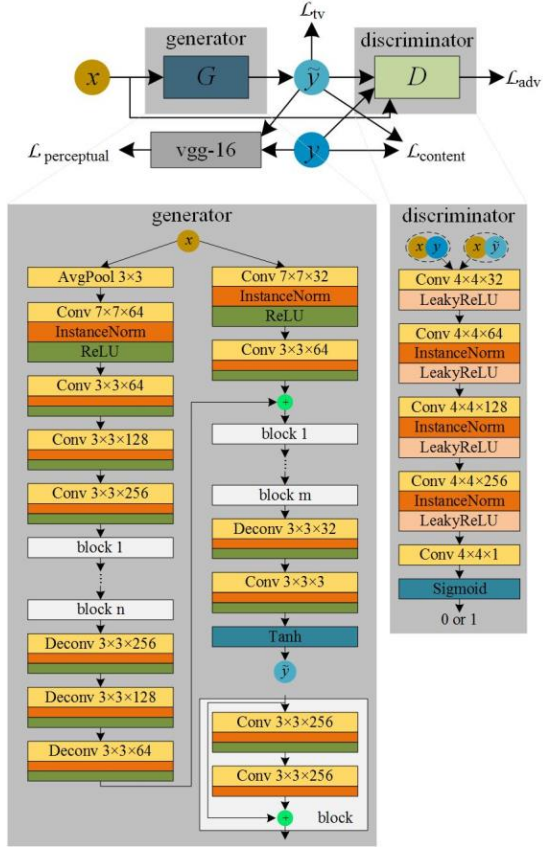

Figure 2: Proposed TIC-CGAN architecture.

### 3.1. Content loss

The pixel-wise MSE loss is widely used to encourage the network output to be close to the ground-truth label image. We use the pixel-wise MSE loss to minimize the low-level content errors like global chrominance and luminance between colorized thermal and corresponding RGB images. The low-level content loss is then defined as:

$$\mathcal{L}_{\text{content}} = \frac{1}{WH} \sum_{i=1}^{W} \sum_{j=1}^{H} \left\| y_{i,j} - G(x)_{i,j} \right\|_1, \quad (1)$$

where $W$ and $H$ denote the height and width of the thermal image respectively, and $\left\| \cdot \right\|_1$ denotes the L1 norm.

### 3.2. Adversarial loss

While using content loss favors high PSNR, the colorized results will lose the detail content that can be seen in [12]. To encourage the network to output colorized results with more realistic details, an adversarial loss is adopted. The adversarial loss is used to make colorized results indistinguishable from real RGB images and is defined as:

$$\mathcal{L}_{\text{adv}} = \mathbb{E}_x[-\log(D(G(x), x))], \quad (2)$$

where $x$ is not only the input of the generator but also the input of the discriminator as a conditional item [14].

### 3.3. Perceptual loss

It is well known that using adversarial loss in image synthesis tasks are prone to produce distorted textures [13], which also occurs in our thermal image colorization task. To alleviate this problem, we use a perceptual loss based on ReLU activations of the VGG-16 network, following [29]. The perceptual loss is beneficial to keep textures in the high-level space and is defined as:

$$\mathcal{L}_{\text{perceptual}} = \sum_k \frac{1}{C_k H_k W_k} \sum_{i=1}^{H_k} \sum_{j=1}^{W_k} \left\| \phi_k(y)_{i,j} - \phi_k(G(x))_{i,j} \right\|_1, \quad (3)$$

where $\phi_k(\cdot)$ represents the feature representations of the kth maxpooling layer in the VGG-16 network, and $C_k H_k W_k$ represents the size of these feature representations.

### 3.4. TV loss

In addition to using perceptual loss to recover high-level content, we further adopt a TV loss [38] to enhance the spatial smoothness of the colorized thermal images and the TV loss is defined as:

$$\mathcal{L}_{\text{tv}} = \frac{1}{WH} \sum \left| \nabla_x G(\tilde{y}) + \nabla_y G(\tilde{y}) \right|, \quad (4)$$

where $|\cdot|$ denotes the element-wise absolute value of the given input.

### 3.5. Full objective

The final TIC-CGAN loss that optimizes network parameters in $G$ is defined as:

$$\mathcal{L}_{\text{total}} = \mathcal{L}_{\text{content}} + \lambda_{\text{perceptual}} \mathcal{L}_{\text{perceptual}} + \lambda_{\text{adv}} \mathcal{L}_{\text{adv}} + \lambda_{\text{tv}} \mathcal{L}_{\text{tv}}, \quad (5)$$

where $\lambda_{\text{perceptual}}$, $\lambda_{\text{adv}}$ and $\lambda_{\text{tv}}$ represent the weights that control the share of different losses in the full objective function respectively. The setting of weights is based on preliminary experiments on the training dataset.

| Method | PSNR | NQM | SSIM | MSSIM |
|---|---|---|---|---|
| Naïve | 9.15 | -1.09 | 0.32 | 0.13 |
| TIR2Lab | 14.6 | 3.40 | 0.52 | 0.51 |
| TIC-CGAN | **15.85** | **4.44** | **0.53** | **0.52** |

Table 1: Average results of 29179 images in the test dataset. Best results are in **bold.**

### 3.6. Network architecture

Figure 2 illustrates the overall framework of our system. Adapted from [13], TIC-CGAN has the generative network composed of two sub-networks. Each sub-network architecture follows the one proposed by Johnson *et al.* [29]. We use $70 \times 70$ PatchGAN [14] for our discriminative network, which distinguishes whether local image patches are real or fake.

## 4. Experiments

To evaluate the ability and quality of our TIC-CGAN method, we conduct a series of experiments on a dataset containing campus, road and downtown scenes. We also compare against a representative visible grayscale colorization baseline [14] and the latest state-of-the-art thermal infrared colorization baseline by Berg *et al.* [12]. We start our experiments by providing a quantitative evaluation against leading approaches in Sec. 4.1. We then present a subjective evaluation in Sec. 4.2. Finally, we analyze the limitations of our TIC-CGAN approach in Sec. 4.3.

**Implementation details** We use regular GANs [11] rather than LSGANs [39] to define the adversarial loss as LSGANs cannot generate fine textures (see Sec. 4.2 for details). The number of filters in the first convolutional layer in the generator and discriminator is set to 32. The number of blocks in the generator is set to $m = 3$ and $n = 9$, and the weights of different losses in the full objective function are set to $\lambda_{adv} = 0.03$, $\lambda_{perceptual} = 1$ and $\lambda_{tv} = 1$ respectively. The stride of all convolutional layers remains the same as in [13]. We compute the perceptual loss at feature representations $\phi_3$, $\phi_8$, $\phi_{15}$ and $\phi_{22}$. For optimization we use ADAM [40]. We jointly train all networks together on a NVIDIA GTX1080 GPU for 10 epochs with a batch size of 1. The entire training time takes 12h.

**Dataset** We conduct extensive experiments on the KAIST multispectral pedestrian dataset [41] that contains 95k, day and night color-thermal image pairs. Considering that the human eye prefers day images, we only use the day color-thermal image pairs for our thermal infrared colorization task. 33399 color-thermal image pairs in their day training dataset are used for training and 29179 color-thermal image pairs in their day test dataset are used for evaluation. These thermal infrared images are captured using a FLIR A35 microbolometer LWIR camera with a resolution of $320 \times 256$ and upsampled to $640 \times 512$ during image alignment. To reduce memory usage and training time, we resize the resolution of all images to $320 \times 256$, following [12].

**Baselines** We compare our approach with two state-of-the-art methods: Naïve and TIR2Lab. Naïve is trained to estimate the chrominance from the luminance of the thermal infrared image itself using an existing visible grayscale colorization approach. Unlike [12], we choose the approach proposed by Iosla *et al.* [14] as Naïve, which leverages conditional adversarial networks for various image-to-image translation tasks and also shows excellent performance in colorizing grayscale visible images. The TIR2Lab proposed by Berg et al. [12] is trained to estimate both the luminance and the chrominance of thermal infrared images. We train Naïve from scratch on our dataset. The implementation of TIR2Lab is from the publicly available codes provided by the authors.

### 4.1. Quantitative evaluation

We first quantitatively compare Naïve [14], TIR2Lab [12] and our TIC-CGAN methods on the task of transforming thermal infrared images to RGB images. The obtained average scores on the test dataset are reported in Table 1. Peak signal-to-noise ratio (PSNR), noise quality measure (NQM) [42], structure similarity index (SSIM) [43], and multiscale structure similarity index (MSSIM) [44] are used as the quantitative evaluation metrics (the higher, the better). PSNR estimates absolute errors, NQM, SSIM and MSSIM measure the perceived quality, surpassing PSNR. From Table 1, we can clearly find that TIC-CGAN outperforms Naïve and TIR2Lab in all measurement metrics. This shows that our method performs well both in terms of global quality and local perception. The following experiments will confirm this visually.

### 4.2. Qualitative evaluation

We further compare our method via a subjective evaluation. Figure 3 provides six colorized examples using the Naïve, TIR2Lab and TIC-CGAN approaches. From Fig. 3b, we can see that the performance of the Naïve approach is unstable. Although Naive successfully estimates the chrominance in some cases (ex. 3, 5, 6), it produces chaotic colors (ex. 1, 4) or even no colors in others (ex. 2). In addition, since the luminance is taken directly from thermal infrared images, the colorized results by Naïve still look unrealistic. TIR2Lab roughly recovers the chrominance as a whole, but fails to accurately estimate the luminance as shown in Fig. 3c. The cars (ex. 4), buildings (ex. 3, 4) and trees (ex. 1, 2, 3, 5, 6) colorized by the TIR2Lab approach are heavily blurred and lack fine textures. The main reason for this problem is that their objective function is only defined by the L1 norm, which only computes global errors and ignores local changes. Further, in some cases (ex. 4) TIR2Lab produces artifacts in the colorized sky. In contrast, using the proposed composite objective function, our TIC-CGAN can accurately generate local realistic textures and fine details, and achieve more plausible RGB images in all cases as shown in Fig. 3d.

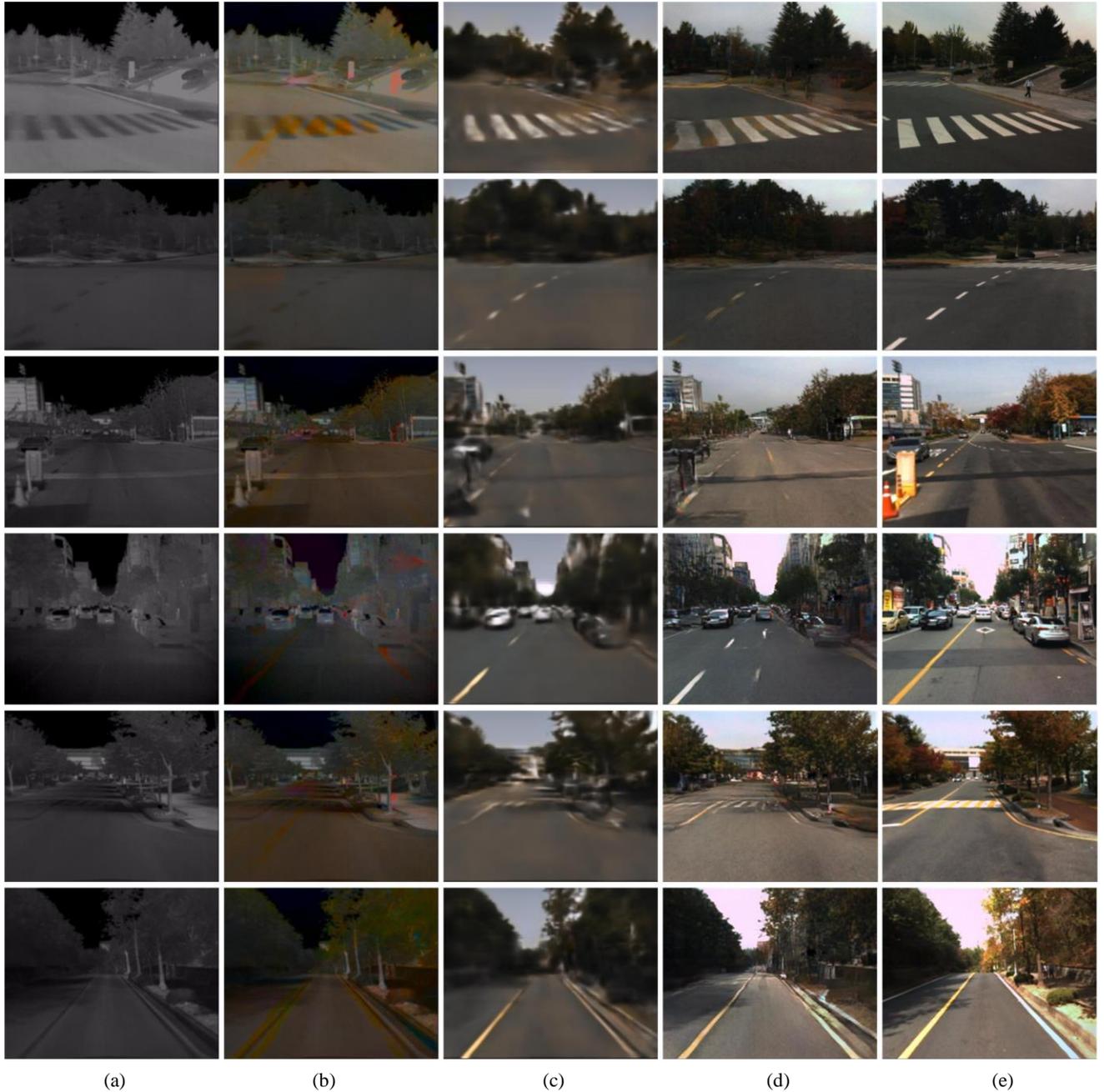

(a) (b) (c) (d) (e)

Figure 3: Colorized results using three different methods. (a) Thermal infrared images. (b) Naïve. (c) TIR2Lab. (d) TIC-CGAN. (e) True RGB images.

**Night to day** We also compare against the TIR2Lab approach on the experiment of colorizing thermal infrared images captured at night. Figure 4 shows five colorized examples using the TIR2Lab and TIC-CGAN methods that are both trained on day images. Due to the low ambient temperature at night, the contrast of night thermal infrared images is lower than that of day thermal infrared images, increasing the difficulty of recognizing different objects in the scene. So the colorized results on night thermal infrared images using TIR2Lab (see Fig. 4b) look more blurred than on day thermal infrared images (see Fig. 3c). The building, people, cars and trees in the colorized results are almost indistinguishable. In contrast, our TIC-CGAN is still able to recover most details which are clearly visible in Fig. 4c. Indeed, since the difference between night and day thermal infrared images, TIC-CGAN trained on day dataset produces slightly worse results on night thermal infrared images but significantly outperforms TIR2Lab.

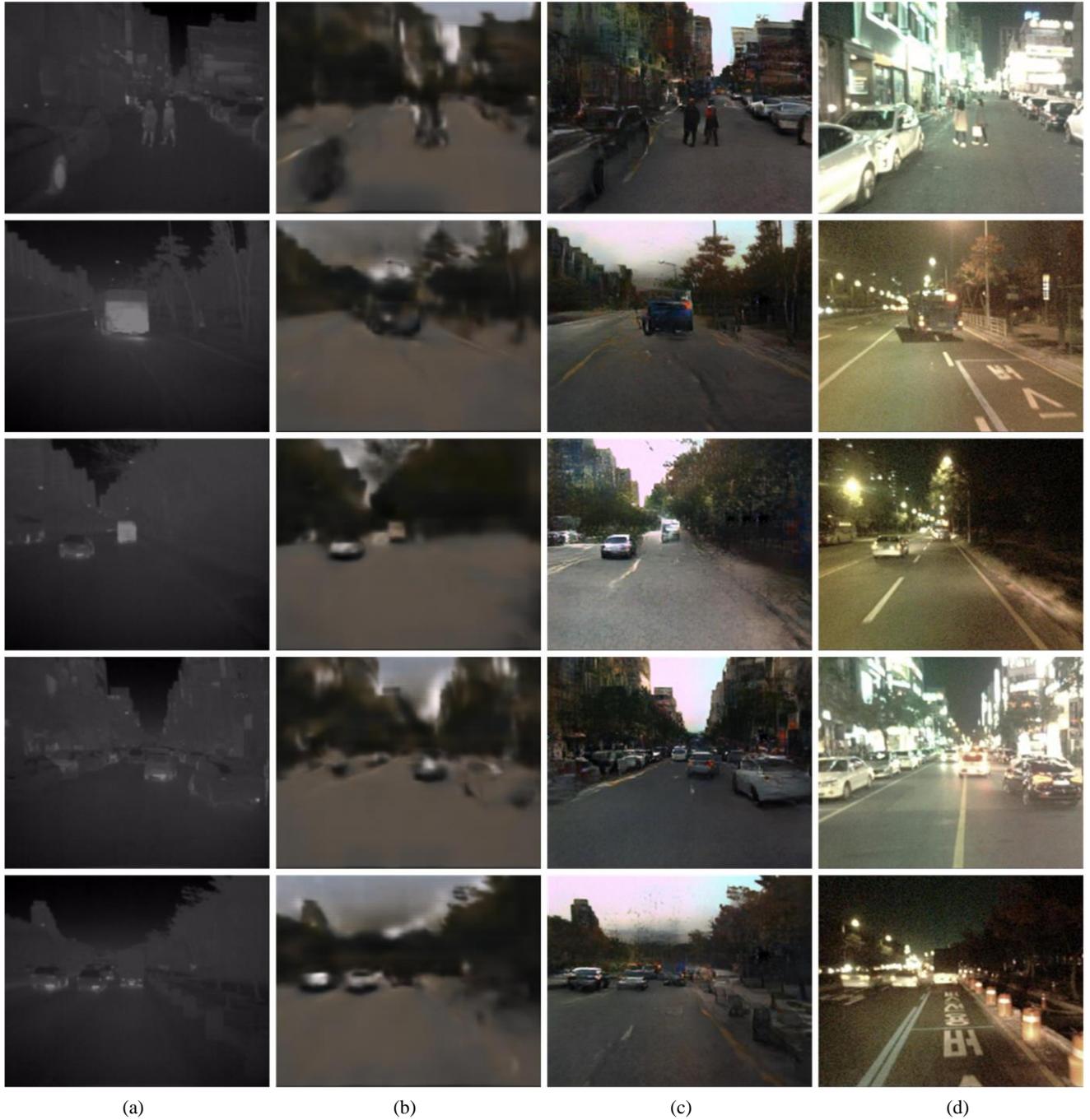

(a) (b) (c) (d)

Figure 4: Colorized results using TIR2Lab and TIC-CGAN. (a) Night thermal infrared images. (b) TIR2Lab. (c) TIC-CGAN. (d) True RGB images.

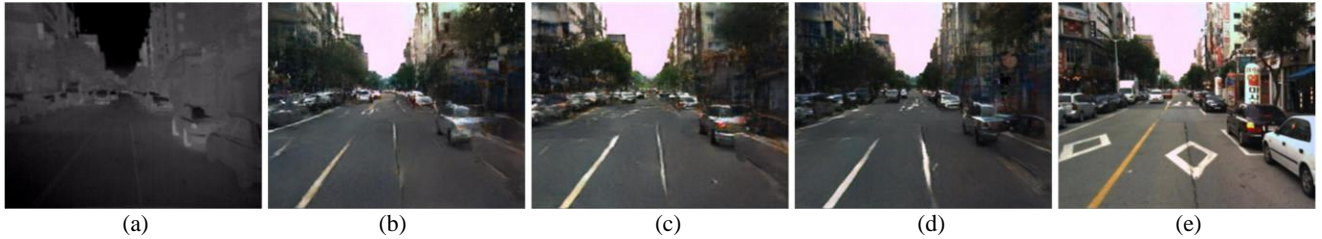

(a) (b) (c) (d) (e)

Figure 5: Colorized results using different generator architectures. (a) Thermal infrared images, (b) UNet, (c) ResNet, (d) Our, (e) True RGB images.

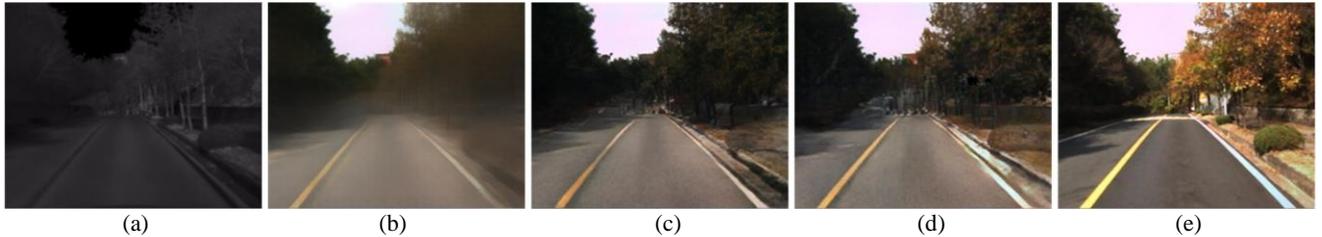

(a) (b) (c) (d) (e)

Figure 6: Colorized results using adversarial loss with different definitions. (a) Thermal infrared images, (b) TIC-LSCGAN, (c) TIC-GAN, (d) TIC-CGAN, (e) True RGB images.

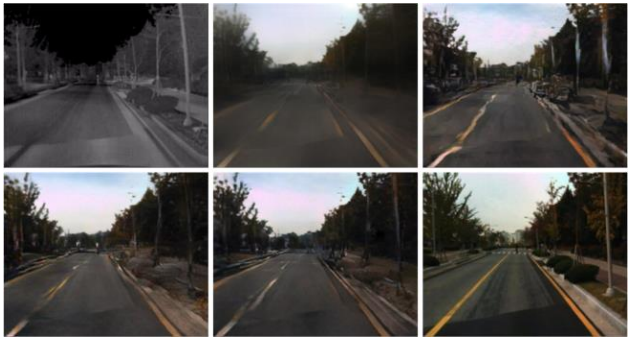

Figure 7: From left to right, top to bottom: thermal infrared images, results by removing adversarial loss, results by removing perceptual loss, results by removing TV loss, results by our default loss, true RGB images.

**Generator architecture** We compare results using different generator architectures with all the other components fixed (Fig. 5). Specifically, we compare against two state-of-the-art generator structures: UNet [14] and ResNet [29]. The number of residual blocks in the ResNet is set to 9. As can be seen, our generator architecture achieves more natural results with sharp and smooth details than other generator architectures. This confirms that the coarse-to-fine generator specially designed for high-resolution image synthesis task [13] is the most suitable network for thermal infrared colorization.

**Definition of the adversarial loss** We study the definition of the adversarial loss with two different definitions: TIC-GAN and TIC-LSCGAN. TIC-GAN is to remove the condition term in the adversarial loss and TIC-LSCGAN is to use LSGANs [39] to compute the adversarial loss. All other parameter settings remain the same. We can clearly find that using LSGANs produces very blurred results in Fig. 6b. The main reason is that, similar to PSNR, LSGANs mainly compute global error and ignore local perception. In addition, adding the condition term in the discriminator is beneficial for generating finer textures (Fig. 6c, 6d).

**Ablations of the full loss** We also explore the impact of each loss in our objective function. Comparison results are shown in Fig. 7. As can be seen, removing the adversarial loss cannot generate details accurately, and removing the perceptual loss leads to distorted details. Removing TV loss seems to produce the same results as our full loss, but the details are less smooth. Therefore, each loss is indispensable for generating high-quality colorized thermal infrared images.

### 4.3. Limitations

Although the proposed approach has achieved impressive results in most cases, some defects are inevitable. Figure 8 shows four failure cases. Our approach often succeeds in colorizing thermal infrared images captured when the car goes straight (*e.g.*, see Figs. 3d, 5d, 6d). However, our approach produces very poor results on thermal infrared images taken when the car is turning, where the details are seriously distorted and blurred (see Fig. 8b, 8c, 8d). This might be caused by the incomplete training dataset, where few training images are captured when the car turns. This phenomenon also occurs in colorizing images taken at the campus entrance (see Fig. 8a). In addition, since colorization does not have a unique solution, there may be multiple colors for the same object in two adjacent frames in the thermal infrared video. Therefore, although the visual effect on a single frame image is satisfactory, colorized images in the video looks incoherent. This video is available in the supplementary material.

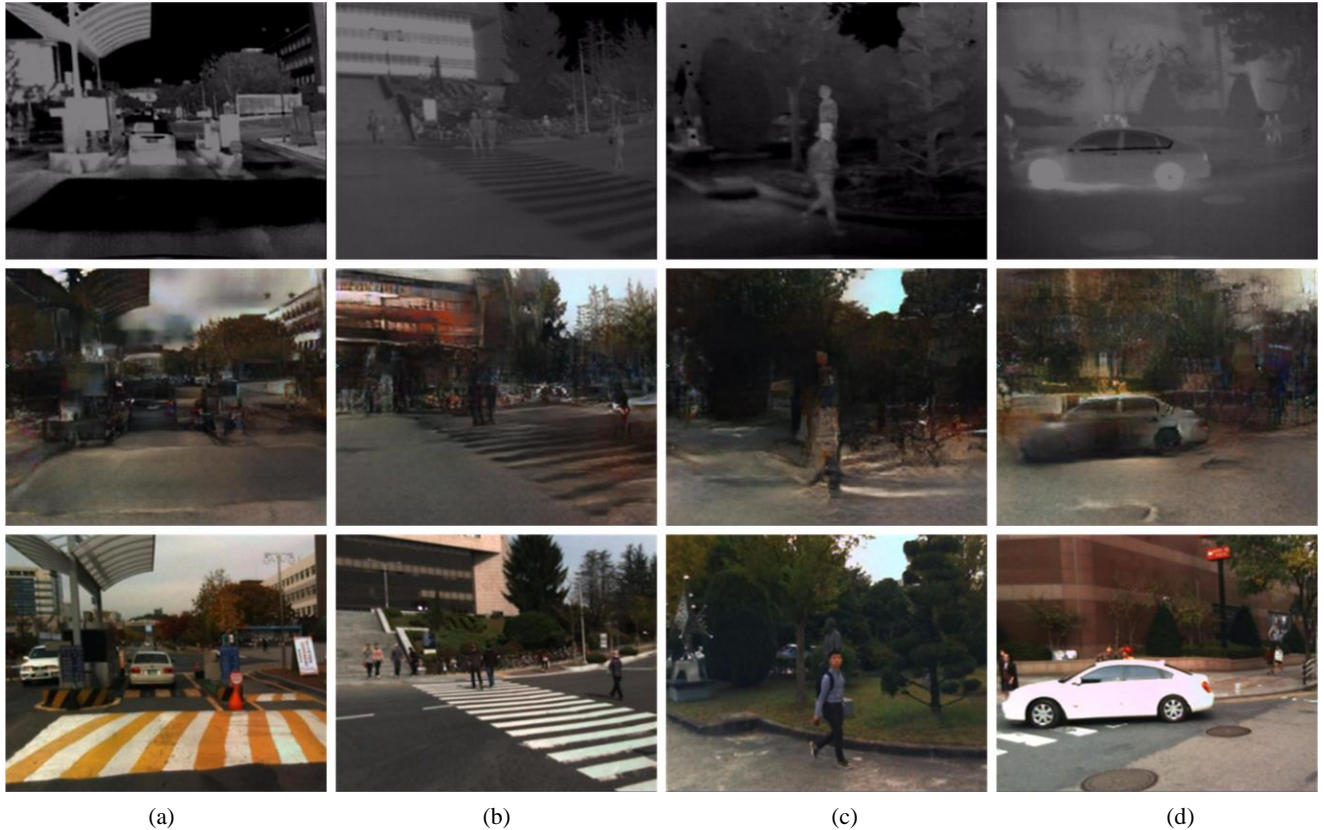

|   (a)   |   (b)   |   (c)   |   (d)   |

Figure 8: Failure colorized cases for our approach. From top to bottom: thermal infrared images, results from TIC-CGAN, true RGB images.

## 5. Conclusions

In this work, we proposed TIC-CGAN – a conditional generative adversarial network to generate RGB images from thermal infrared images. In contrast to the previously proposed approach [12] that only restored rough luminance and chrominance information, this method uses a coarse-to-fine generator and a composite objective function that combines content, adversarial, perceptual and TV losses to produce results with realistic colors and fine details. Quantitative and qualitative assessments reveal that our method outperforms the state-of-the-art deep learning-based approaches [12, 14].

Further work is twofold. First, in order to build a complete training dataset, collecting more thermal infrared to RGB images is needed. Second, exploring more different generator architectures is needed to further improve the image quality.